\documentclass[runningheads]{llncs}
\usepackage{graphicx}

\usepackage{optidef}
\usepackage{authblk}
\usepackage{cite}
\usepackage[dvipsnames]{xcolor}
\usepackage{amssymb}
\usepackage{romannum}
\usepackage{amsfonts}
\usepackage{pifont}

\usepackage{enumitem}
\usepackage{path}
\usepackage{verbatim}
\usepackage{algorithm}
\usepackage{algorithmic}
\usepackage{framed}
\usepackage{footnote}
\usepackage{color}
\usepackage{amsmath}
\usepackage{cases}
\usepackage{theorem}
\usepackage{xurl}
\usepackage{tabularx}
\usepackage{booktabs}
\usepackage[colorlinks=true, allcolors=black]{hyperref}
\usepackage{threeparttable}
\usepackage{graphicx}
\usepackage{pgfplots}
\usepackage{pgfplotstable}
\pgfplotsset{compat=newest}
\graphicspath{{graphics/}}
\usepackage{tablefootnote}
\usepackage{subcaption}

\begin{document}
\title{Optimal Design and Implementation of an Open-source Emulation Platform for User-Centric Shared E-mobility Services}
%
\titlerunning{User-Centric Shared E-mobility Platform}
\author{Maqsood H. Shah \and 
 Yue Ding \thanks{Joint first author}\and 
 Shaoshu Zhu \and 
 Yingqi Gu \and 
 Mingming Liu}

\institute{
Insight SFI Research Centre for Data Analytics, Dublin City University
\email{mingming.liu@dcu.ie}}


\authorrunning{M.H. Shah et al.}


%
\maketitle              
\begin{abstract}
With the rising concern over transportation emissions and pollution on a global scale,  shared electric mobility services like E-cars, E-bikes, and E-scooters have emerged as promising solutions to mitigate these pressing challenges. However, existing shared E-mobility services exhibit critical design deficiencies, including insufficient service integration, imprecise energy consumption forecasting, limited scalability and geographical coverage, and a notable absence of a user-centric perspective, particularly in the context of multi-modal transportation. More importantly, there is no consolidated open-source platform which could benefit the E-mobility research community. This paper aims to bridge this gap by providing an open-source platform for shared E-mobility. The proposed platform, with an agent-in-the-loop approach and modular architecture, is tailored to diverse user preferences and offers enhanced customization. We demonstrate the viability of this platform by providing a comprehensive analysis for integrated multi-modal route-optimization in diverse scenarios of energy availability, user preferences and E-mobility tools placement for which we use modified Ant Colony Optimization algorithm so called Multi-Model Energy Constrained ACO (MMEC-ACO) and Q-Learning algorithms. Our findings demonstrate that Q-learning achieves significantly better performance in terms of travel time cost for more than 90\% of the instances as compared to MMEC-ACO for different scenarios including energy availability, user preference and E-mobility tools distribution. For a fixed (O, D) pair, the average execution time to achieve optimal time cost solution for MMEC-ACO is less than 2 seconds, while Q-learning reaches an optimal time cost in 20 seconds on average. For a run-time of 2 seconds, Q-learning still achieves a better optimal time cost with a 20\% reduction over MMEC-ACO's time cost. 

\end{abstract}

\section{Introduction} 
	\label{introduction}

Environmental degradation, significantly driven by the transportation sector, is a pressing global challenge. One plausible solution to deal with it is to promote the deployment of shared E-mobility services. By integrating various E-mobility tools, such as E-scooters, E-bikes, and E-cars, into the overall transportation ecosystem, we can maximize the benefits of these alternatives in an integrated and user-centric manner
\cite{dias2021therole,nikitas2021cycling}.  This shared E-transportation paradigm not only offers users a new way of traveling but also plays a crucial role in reducing car ownership and emissions during operation. As a result, it has become an essential part of the Mobility as a Service (MaaS) paradigm in many cities across the globe \cite{liao2022electric}. However, as many shared E-mobility services roll out into the market, many challenges are anticipated. For instance, efficient integrability, scalability, user-centric design, and real-time data analytics are a few key aspects. Some recent works in this regard, including \cite{liao2022electric,duraismay2021adpaptive,chen2021review}, have also highlighted these aspects as some of the critical gaps in the existing E-mobility research.  To overcome these challenges, it is imperative to have an integrated platform that can emulate different traffic scenarios, manage the number and location of charging stations/E-hubs, and provide operators with real-time analytics. 

Recent studies in the context of E-mobility have focused on several disparate aspects therein. For instance, in \cite{cho2022digital} and \cite{ferrara2019simulation}, the authors explored the integration of E-mobility with a particular focus on energy distribution systems and energy management, respectively. In \cite{echternacht2018simulating},  the authors used a data-driven simulation approach to assess the impact of growing numbers of Electric Vehicle (EV) users and private charging stations. Similarly, in \cite{campisi2022promotion,laurischkat2016business}, authors explored the impact of shared E-mobility on urban transportation and infrastructure distribution. In slightly more related works \cite{luo2021deployment,luo2023fleet},  authors introduced a multi-agent simulation platform based on deep reinforcement learning optimization. However, this platform particularly focused on addressing the challenge of fleet rebalancing, which is not the main focus of our paper. Furthermore, in \cite{buchmann2021stimulating}, authors explored E-mobility diffusion in Germany using an agent-based simulation, focusing specifically on electric cars. 
The \textit{Flow} architecture proposed in \cite{wu2017flow} provides a reinforcement learning framework for  benchmarking in traffic control. 
Our work distinguishes itself by offering a generic, open-source platform that is user-centric and accommodates micromobility options as well.
Keeping aforementioned aspects in view, and in order for us to have a user-centric solution that encourages widespread adoption of E-mobility services for better impact, an open-source emulation platform is highly essential. This would ensure the development of robust solutions and will serve as a practical resource for stakeholders to collectively advance and refine E-mobility research. 

Before summarizing the major contributions of our work, it is also essential to point out some existing commercially available platforms for shared E-mobility such as Moovit \cite{moovit2023}, Bird \cite{bird2023}, ElectricFeel \cite{electricfeel2023} and Wunder Mobility \cite{wondermobility2023}. Many commercial platforms in the E-mobility sector, typically operated by either regional or multinational startup ventures, often lack transparency about system architectures and algorithms, which complicates comprehensive research and analysis. Also, their reliance on real-world setups restricts their capability to understand various scenarios, especially if they are deployed at large-scale.

This paper presents a scope of work distinct from existing commercial platforms. Specifically, we focus on proposing an open-source platform which is able to address the multi-modal routing optimization problem in a user-centric fashion. To the best of our knowledge, there is still a lack of consolidated, open-source user-centric platform in the context of shared E-mobility. This presents a considerable challenge for the broader research community in E-mobility. Our work aims to fill this gap by offering a collaborative and transparent platfrom that addresses crucial challenges in the E-mobility research domain. In summary, the major contributions presented in this paper are as follow: 
\begin{itemize}
    \item We propose an open-source platform\footnote{The accompanying code for this paper is available at \url{https://github.com/SFIEssential/Essential}} with ability to emulate different traffic and deployment scenarios, integrating user-preferences, energy consumption prediction and has the ability to provide multi-modal routing optimization. The platform aims to advance the state of E-mobility research by enabling exploration and analysis of diverse transportation scenarios, optimization constraints and integration of various E-mobility resources.
    \item We formulate the multi-modal routing problem as a constrained shortest path problem that incorporates energy constraints, user preferences, and multi-modality. To address this challenge, we have developed two specialized algorithms: a modified meta-heuristic Ant-Colony Optimization algorithm tailored to our operational setup, named Multi-Modal Energy-Constrained Ant-Colony Optimization (MMEC-ACO), and the Q-learning algorithm, adapted for the same problem setting.
    \item We evaluate the proposed platform using a range of use case scenarios, specifically focusing on Dublin City Centre. This includes a comparative analysis of MMEC-ACO and Q-learning to assess the optimality (travel time cost) and scalability (execution time) of both algorithms under varying conditions such as energy levels, E-hub placements, and user preferences. This analysis is aimed at determining the suitability of each algorithm across multiple scenarios, providing valuable insights into their performance and scalability.
\end{itemize}

The rest of the paper is organized as follows. In Section \ref{PA}, we provide a detailed overview of the proposed architecture. The methodology Section, encompassing problem setup, formulation, and optimization algorithms, is discussed in Section \ref{OPnEC}. The evaluation of optimization methods with experimental statement, results and discussion is entailed in Section \ref{sec:exp_results}. Finally, we conclude the paper and outline scope for future research in Section \ref{conc}.
\section{Platform Architecture} \label{PA}
        
        \begin{figure*}[ht]
        \centering
        \includegraphics[width=\textwidth,height=6cm]{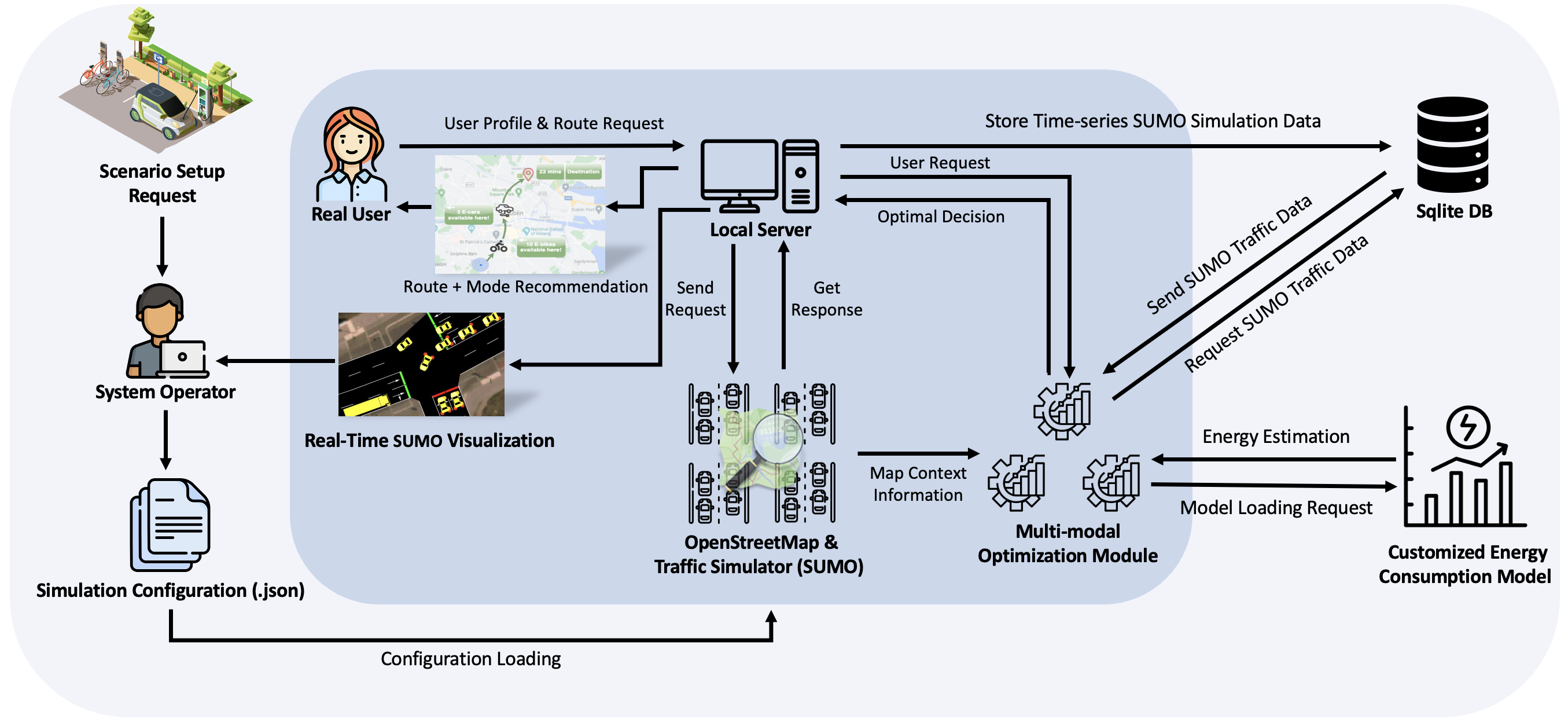}
        \caption{System Architecture Diagram}
        \label{fig:figure1}
        \end{figure*}

       Our proposed system architecture is shown in Fig. \ref{fig:figure1}. The system consists of several functional blocks designed to recommend a personalized multi-modal route based on a real user's input within a specific scenario defined by the system operator. Specifically, the system consists of  simulation, optimization, and visualization modules, and based on the architecture, it operates as follows: The platform integrates a simulator, enhanced with custom plugins to represent varied traffic and E-mobility scenarios. It employs geographical data from OpenStreetMap \cite{haklay2008openstreetmap} for realistic simulations of environments like E-hubs and EVs. The system updates and displays these scenarios in real-time, and the optimization module processes this data to provide dynamic multi-modal route recommendations requested by users, which are visually presented for an interactive user experience. Additionally, it integrates an energy consumption prediction model that provides personalized forecasts for individual users. In the following, we outline details of each functional block.

        \subsection{Simulation}
       The simulation module within the proposed architecture encompasses multiple dimensions, with traffic simulation standing out as a crucial component in the overall design task. We had several options to choose from the available repositories of available open-source traffic simulation tools \cite{barcelo2005dynamic,fellendorf2010microscopic}. We chose Simulation for Urban Mobility (SUMO) \cite{krajzewicz2002sumo} due to its exceptional flexibility, easy portability, and scalability. SUMO empowers us to create customized simulations tailored to the complexities of urban networks. 

        As illustrated in the Fig. \ref{fig:figure1}, the simulation process begins with a scenario setup request initiated by a system operator who configures the simulation environment using a JSON file, defining simulation parameters and loading configuration data. Once setup is complete, the system operator can visualize the real-time SUMO simulation, which interacts closely with several core components. A local server plays a central role, receiving user profiles and route requests, and subsequently sending these requests to the multi-modal optimization module. This module, based on information from the OpenStreetMap and SUMO, generates optimal route and mode recommendations. 
        The system leverages real-time SUMO visualization and stores traffic data in an SQLite database, allowing for the retrieval and analysis of historical traffic patterns, which is crucial for optimizing future route recommendations. 
 \subsection{Optimization}
        Optimization is one of the core modules of our platform. It is designed to facilitate personalized route planning from end-user perspective and has the ability to offer tailored recommendations based on the user preferences, energy availability and other implementation constraints. The platform has the flexibility to integrate custom optimization algorithms for additional case studies and future research by leveraging the simulation data. More details pertaining to the optimization algorithms that we use for evaluation are provided in Section \ref{OPnEC}.

        \subsection{Visualization}
\begin{figure*}[!ht]
        \centering
        \includegraphics[width=0.9\textwidth,height=6cm]{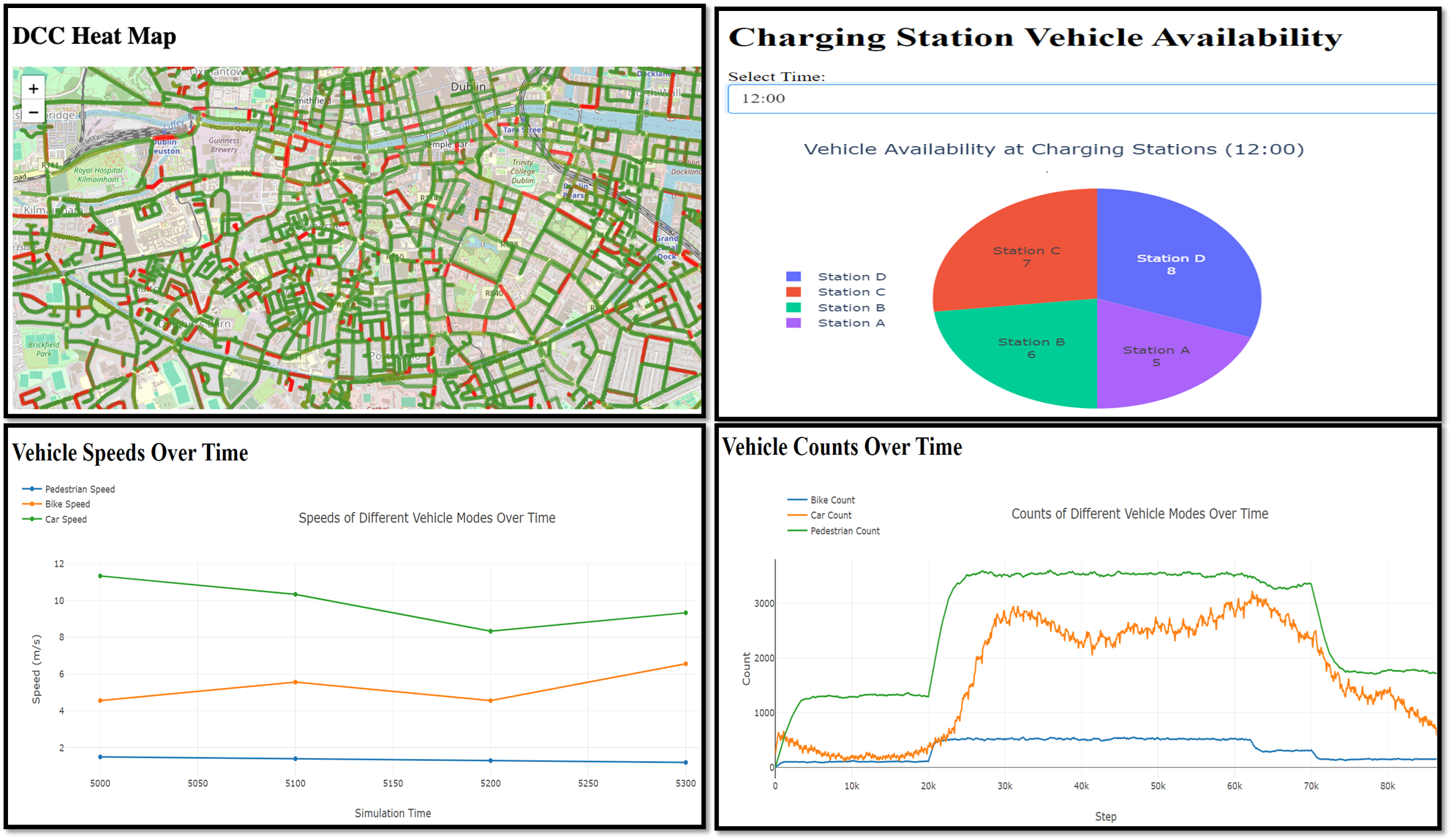}
        \caption{Different interfaces to illustrate the platform's visualisation module}
        \label{fig:app}
        \end{figure*}

        The main purpose of the visualization module is to present data in a visual format that is easily understandable and interpretable. It offers an interactive experience that allows users to actively engage with and explore the simulated environment. Through intuitive controls and real-time feedback, users can manipulate variables, observe the effects, and simulate ``what-if" scenarios. 
        Some of the visualisations pertaining to  different aspects at simulation stage such as, the number of vehicles on the road, average speed of different E-mobility types and availability of E-mobility types on different E-hubs are illustrated in the Fig. \ref{fig:app}. The Dublin City Centre (DCC) map shown on the top left corner indicates the heat map based on the congestion level along different segments of the map which varies for various E-mobility types and times of day depending on simulated scenarios and the traffic patterns. From the end-users' perspective, an interactive and personalized route recommendation app and web-based application is also designed which could not be shown here due to space limitation\footnote{Readers can see the app layout in the project's github repo: \url{https://github.com/SFIEssential/Essential}}.
         
        \subsection{Energy Consumption Model}
        Energy consumption model is another important aspect of our platform which directly addresses ``range anxiety" (being one of the primary limiting factors in E-mobility). By incorporating energy consumption of each E-mobility tool as part of the optimization problem, we aim to alleviate the user's concern about the range anxiety. We utilize several models including mathematical model \cite{Burani2022} and multiple data-driven models to predict the energy consumption for each E-mobility device along the path. For more details pertaining to our work on the energy consumption modeling, readers are encouraged to refer \cite{ding2024datadriven,Sen_Fed,energy_review}.


\section{Methodology}\label{OPnEC}
\begin{figure}[ht]
        \centering
        \includegraphics[width=0.8\linewidth, height = 4.8cm ]{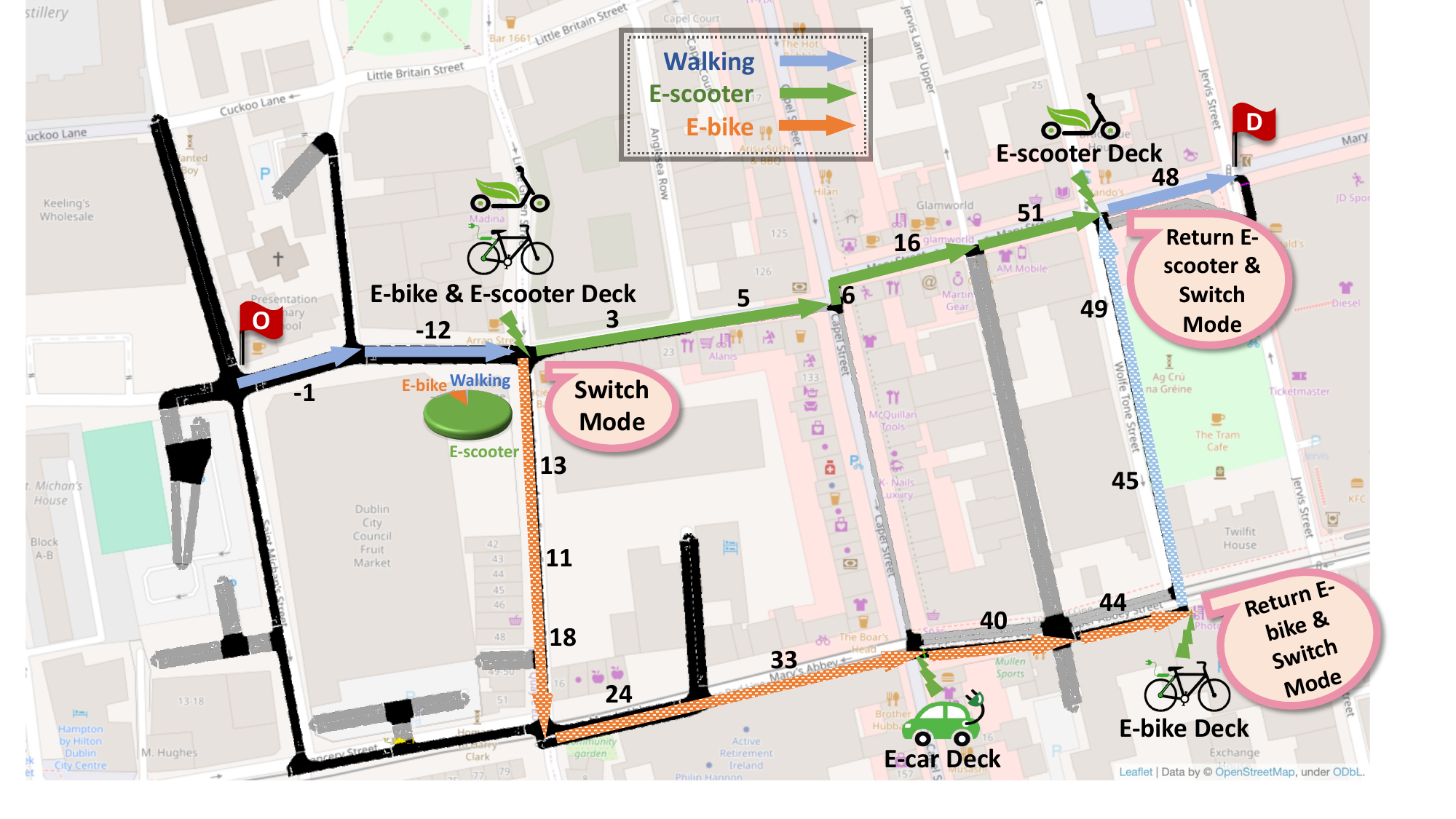}
        \caption{Illustration of the Multi-Modal Route Optimization Scenario highlighting the E-mobility docking stations and possible multi-modal routes}
        \label{fig:figure4}
\end{figure}

The problem at hand is to recommend an optimal path that minimizes the total travel time cost from an origin (O) to a destination (D) using various E-mobility options, considering various constraints based on user preferences, energy availability, and multi-modality. Imagine a scenario illustrated in Fig. \ref{fig:figure4}, where a user has to reach a destination point (marked with flag `D') from their current location (marked as flag `O') using E-mobility options. We designate specific locations (called E-hubs) for various types of E-mobility, as depicted in Fig. \ref{fig:figure4}. The markings on the map (e.g., -1, -12, etc.,) correspond to the edge IDs obtained from the simulation module which are used to generate the graph. This simple illustration shows two different possible options to travel from ``O" to ``D", whereas in reality the actual number of possible options will be much higher (depending on the actual map size, number of E-hubs/charging stations, deployed E-mobility tools, energy availability which are provided by the simulation module).

In simplest terms, the primary objective is to recommend the optimal path with minimal travel time cost based on a set of constraints. To achieve the optimal solution for above mentioned problem, we use metaheuristic based ACO and reinforcement learning (Q-learning) approaches, which are more suitable for constrained shortest path problems due to the following reasons: To begin with, classic shortest path algorithms, such as Dijkstra’s algorithm \cite{Dijkstra1959}, are not suitable for our problem as they do not adequately address the energy constraints required for our system setup. Traditional mathematical programming methods, such as Integer Linear Programming (ILP), may handle certain types of constrained shortest path problems demonstrated in paper \cite{DEANDRADE2018141}. However, we prefer heuristic and learning-based methods like Ant Colony Optimization (ACO) and Q-learning because they more accurately model energy dependencies and handle uncertainties. These algorithms track the energy use of an E-mobility tool moving through multiple nodes without needing vehicle changes at each stop, enhancing flexibility in exploring search spaces, which is often challenging to incorporate in traditional methods \cite{Musa2023}. In addition, heuristic and learning-based methods are better suited to handle dynamic and uncertain environments as in our system \cite{drones6110365}, making them more robust and adaptable to real-world applications. Finally, these methods are often designed to provide solutions more quickly than exact methods, which is crucial for real-time applications where timely decision-making is favourable. The details of the problem formulation is explained in the following sections.

\subsection{Problem Formulation}

Through a real-time communication between different modules including simulation and energy consumption modeling we utilize the map data files (.net and .xml) specific to the Region of Interest (ROI) from SUMO, along with various dynamic and static parameters, e.g., average speed and location information for the e-hubs. We translate the given problem into a graph which serves as the foundational structure for the implementation of optimization algorithm for obtaining the optimal route. Our graph modeling method is detailed below.

\subsubsection{Graph Modelling:}
We model the ROI using a weighted directed graph $G = (V, E, W)$, where $V = \{1, \ldots, n\}$ is the vertex set representing road segments or nodes, and $E = \{e_1, \ldots, e_m\}$ is the set of edges with corresponding positive weights in $W^{s} = \{w_1^{s}, \ldots, w_m^{s}\}$. Here, superscript $s$ corresponds to the available E-mobility tools (e.g., E-bike, E-car, E-scooter, etc.). The weight $w_i^{s}$ linked to edge $e_i$ denotes the travel time cost when utilizing the $s$-th type of mobility tool on that edge, with $s \in M$ encompassing all available E-mobility tool types in the network. The weights are assigned based on average speeds corresponding to different edges, derived from simulation. 


\noindent\textbf{Remark on Scalability:} To enhance the scalability of the algorithms, we reduce graph complexity by leveraging the fixed positions of the E-hubs. We utilize the specific location data and E-mobility options associated with these hubs to pre-compute the shortest distances between designated points. This allows us to reconstruct the graph with considerably fewer nodes and edges, facilitating more scalable optimal route recommendations. This streamlined approach limits the search area for ants and Q-learning agents, thereby boosting their efficiency and likelihood of converging on solutions more quickly. In the following, all the experiments are conducted using the reduced graph approach.


\subsubsection{Optimization Objectives and Constraints:}
Given the context above, our optimization problem is to find the optimal path and corresponding mode $s$, i.e., $p = \langle v_1^s, v_2^s, \ldots, v_h^s \rangle$ , which consists of a sequence of vertices starting from the source vertex $v_1 \in V$ and terminating at the destination vertex $v_h \in V$, using the mode $s \in M$, that minimizes the sum of the weights (travel time) of its constituent edges. This is subject to three types of constraints: conformity, energy, and user preference. Conformity constraints ensure seamless transitions between different E-mobility options by requiring transfer stations to facilitate docking for both incoming and outgoing tools. Energy constraints monitor each tool's State of Charge (SOC) to guarantee sufficient energy availability before proceeding to the next segment. Lastly, user-preference constraints ensure the use of E-mobility tools to individual user needs. In the following, we present both MMEC-ACO and Q-learning algorithms to address this optimization problem.

\subsection{MMEC-ACO}
ACO is a metaheuristic algorithm \cite{dorigo2006ant,Liu_ACO} that mimics ants' foraging behavior to optimize routes using virtual pheromones, guiding toward efficient solutions. Pheromone \textit{evaporation} and \textit{deposition} adjust for changing conditions and path quality, balancing exploration and exploitation. We have adapted this for multi-modal optimization within specific constraints including the energy factor. The implementation details of the MMEC-ACO algorithm include the calculation of probabilities for route selection based on pheromone levels, and the mechanism for updating these levels to reflect evolving network conditions. The adapted ACO algorithm integrates vehicle energy levels, supports dynamic mode switching, and enhances pheromone updates based on route efficiency (total travel time), optimizing urban routing effectively.
\newline
\textbf{Next Move Probability:} The probability of selecting the next move step is based on the pheromone level, heuristic information, and energy factor associated with that move:
\begin{align}
& P(S_{t+1} = (e_j, s_j) \mid S_{t} = (e_i, s_i)) = \nonumber \\ 
&\frac{(p_{e_i, s_i \rightarrow e_j, s_j})^\alpha \cdot (h_{e_i, s_i \rightarrow e_j, s_j})^\beta \cdot (\text{ef}_{e_i, s_i \rightarrow e_j, s_j})^\gamma}{\sum_{(e_k, s_k) \neq (e_i, s_i)} (p_{e_i, s_i \rightarrow e_k, s_k})^\alpha \cdot (h_{e_i, s_i \rightarrow e_k, s_k})^\beta \cdot (\text{ef}_{e_i, s_i \rightarrow e_k, s_k})^\gamma}
\end{align}
Here, $P(S_{t+1} = (e_j, s_j) \mid S_{t} = (e_i, s_i))$ is the conditional probability of transitioning from edge $e_i$ with mode $s_i$ to edge $e_j$ with mode $s_j$, given that the current mode at edge $e_i$ is $s_i$. The term $p_{e_i, s_i \rightarrow e_j, s_j}$ represents the pheromone level associated with the transition, $h_{e_i, s_i \rightarrow e_j, s_j}$ denotes heuristic information representing the desirability of the transition, and $\text{ef}_{e_i, s_i \rightarrow e_j, s_j}$ is the binary energy factor associated with the remaining energy for traversing the transition, which is if energy conditions for the transition are met and 0 otherwise. The parameters $\alpha, \beta, \gamma$ control the relative importance of pheromone level, heuristic information, and energy factor, respectively. The denominator represents the sum of pheromone- and heuristic-weighted energy factors for all possible transitions, excluding the current mode $s_i$. Initially, all possible moves are calculated. If a move is infeasible due to the current environmental configurations, the pheromone level for that transition is set to zero, effectively eliminating the possibility of that move.

The Pheromone update rule corresponding to 
\textbf{after Evaporation} stage is given in equation (\ref{eq:after_evap}).
\begin{align}
p_{e_i, s_i \rightarrow e_j, s_j} \leftarrow (1 - \rho) p_{e_i, s_i \rightarrow e_j, s_j}
\label{eq:after_evap}
\end{align}
where $p_{e_i, s_i \rightarrow e_j, s_j}$ is the pheromone level associated with the transition, and $\rho$ is the evaporation rate determining the rate at which pheromone levels diminish over time.
The pheromone update rule corresponding to \textbf{after Deposition} is given in equation (\ref{eq:after_deposition}).
\begin{align}
p_{e_i, s_i \rightarrow e_j, s_j} \leftarrow p_{e_i, s_i \rightarrow e_j, s_j} + \frac{Q}{c_{e_i, s_i \rightarrow e_j, s_j}}
\label{eq:after_deposition}
\end{align}
where $p_{e_i, s_i \rightarrow e_j, s_j}$ is the pheromone level associated with the transition, $Q$ is a constant factor influencing the amount of pheromone deposition, and $c_{e_i, s_i \rightarrow e_j, s_j}$ is the time cost associated with the transition.

\subsection{Q-Learning Algorithm}
While framing the problem as Q-learning \cite{Watkins1992}, each \textbf{state} \textit{s} corresponds to a vertex or node in the graph. The choice of transportation modes (including E-bike, E-car, E-scooter, walking) for moving from one location to another signify the \textbf{action} \textit{a}. The reward for each state-action pair is calculated as the negative travel time of the edge between the current state and the next state for a given mode of transportation in the graph. If the current energy is less than zero, the reward is set to a large negative value to represent a significantly unfavorable outcome. The objective is to minimize the total travel time or cost from the start location to the destination. We start by initializing the Q-table, where each entry \(Q(s, a)\) represents the expected cumulative reward for taking action \textit{a} from state \textit{s}. Then, we iterate until convergence or a maximum number of episodes. We select a state \textit{s} (current location) and choose an action \textit{a} (transportation mode and next node) using an exploration-exploitation strategy (e.g., $\epsilon$-greedy). After taking the action, we observe the reward (negative travel time cost) and the new state \( s' \). Next, we update the Q-value for the current state-action pair using the Q-learning update rule:
\begin{align}
     Q(s, a) = (1 - \alpha) \times Q(s, a) + \alpha \times \left( R(s, a) + \gamma \times \max(Q(s', a')) \right) 
\end{align}
where
 \( \alpha \) is the learning rate,
 \( \gamma \) is the discount factor,
 \( R(s, a) \) is the immediate reward for taking action \( a \) from state \( s \),
 \( \max(Q(s', a')) \) represents the maximum expected cumulative reward for the next state \( s' \). After the Q-table has converged, we can extract the optimal route by choosing the action with the highest Q-value for each state.
\section{Evaluation}\label{sec:exp_results}
To validate the platform and compare the performance of the aforementioned optimization algorithms, we use the actual map of Dublin City Centre \cite{DCC_map_ref}. In the map, nodes represent intersections or junctions where roads meet or end. Edges are the road segments connecting these nodes, defining the paths vehicles can travel. Nodes and edges focus on road network and traffic flow rather than specific buildings or points of interest. This map was chosen as a case study which naturally aligns with the requirement to focus on an Irish location. However, this map also has other advantages, for instance, it has earlier been considered by some other case studies pertaining to the traffic management and hence the simulation in this regard benefits to utilize these traffic patterns \cite{DCC_map_ref}. In addition, with more than 20 thousand edges, the map encapsulates reasonable complexity to validate the scalability aspect for an urban environment. 
\subsection{Experimental Setup} 
Pedestrian and bicycle traffic patterns were created based on information from the 2022 Irish census \cite{CSO2022Table}. The data includes how people in Dublin City commute to work, school, or college, showing that 37,000 people commute by walking and 87,000 by cycling. Over the 24-hour simulation, these numbers were replicated across the map, with peak traffic times starting at 5:30 am and lasting until roughly 7 pm. We define the locations and number of E-hubs based on the traffic patterns data at a given time slot. For the experiments conducted in this paper we use 20 different locations (E-hubs) which support various types of E-mobility tools (E-bike, E-scooter and E-car) with walking as the default option for first and last transitions. The distribution of the tools on these E-hubs can be changed as per the actual demand. It must be noted that we use the reduced graph method briefly discussed Section 3.1 and all the experimental results discussed subsequently are based on it.       
From the algorithms' stand point, we particularly focus on the optimal time cost and execution time for both MMEC-ACO and Q-learning algorithms. We select 500 (Origin, Destination)/(O,D) pairs with varying distances to assess the algorithms' performance in diverse use cases. In particular, we compare the performance of ACO and Q-learning for 1) Initial SOC available for each tool 2) Distribution pattern of the E-mobility tools across E-hubs and 3) impact of changing user preference.  All experiments were performed on a 2022 MacBook Air, equipped with an Apple M2 chip and 16 GB of memory. The operating system used was MacOS version 13.5 (build 2274). The programming environment was Python version 3.11. 
\begin{figure*}[!h]
    \centering
    \begin{subfigure}[t]{0.5\linewidth}
        \centering
        \includegraphics[height=1.5in]{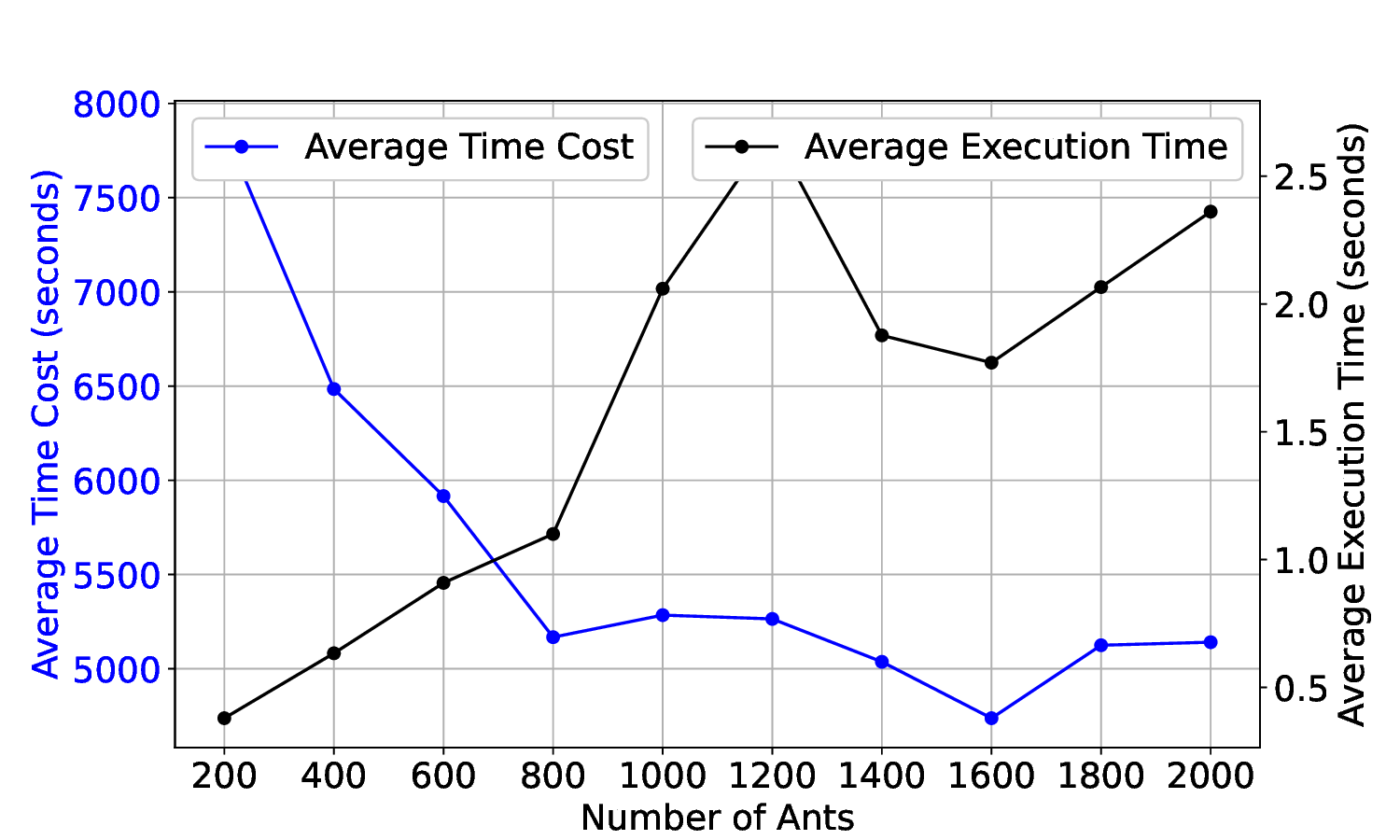}
        \caption{MMEC-ACO}
        \label{subfig:ACO_ants}
    \end{subfigure}%
    ~
    \begin{subfigure}[t]{0.5\linewidth}
        \centering
        \includegraphics[height=1.5in]{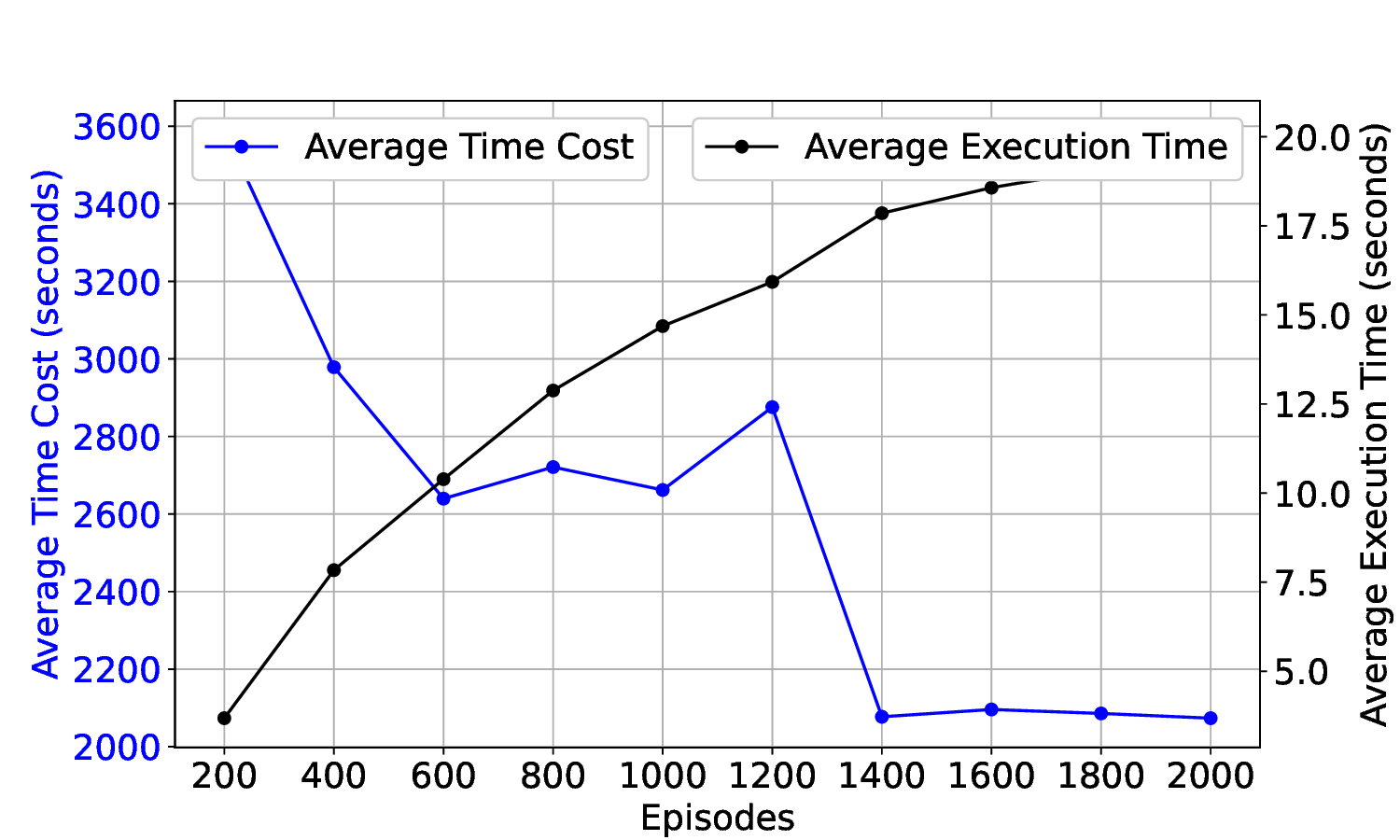}
        \caption{Q-Learning}
        \label{subfig:qlearning_episodes}
    \end{subfigure}
    \caption{Average Execution Times and Travel time Cost for the MMEC-ACO and Q-learning algorithms against different number of Ants and Episodes}
    \label{fig:AllResults}
\end{figure*}
\subsection{Experimental Results}
Before conducting further experiments to evaluate the algorithms' performance under different scenarios, we finalized the primary hyperparameters: the number of ants for MMEC-ACO and the number of episodes for Q-learning. We ran both algorithms with varying numbers of ants and episodes, respectively, using a fixed known (O, D) pair over 30 independent experiments. Figures \ref{subfig:ACO_ants}
and \ref{subfig:qlearning_episodes} present the average time cost (blue, left y-axis) and average execution times (black, right y-axis) for the MMEC-ACO and Q-learning algorithms.
The results indicate that Q-learning generally achieves a better average time cost compared to MMEC-ACO. Specifically, Q-learning reaches an optimal cost of 2100 seconds with an execution time of 20 seconds at 2000 episodes. In contrast, MMEC-ACO on average, achieves an optimal cost of 4500 seconds, but it only takes on average less than 2 seconds to reach this optimal time cost with 1600 ants. 

 \begin{figure}[!h]
        \centering
        \includegraphics[width=0.9\linewidth, height = 6.3cm ]{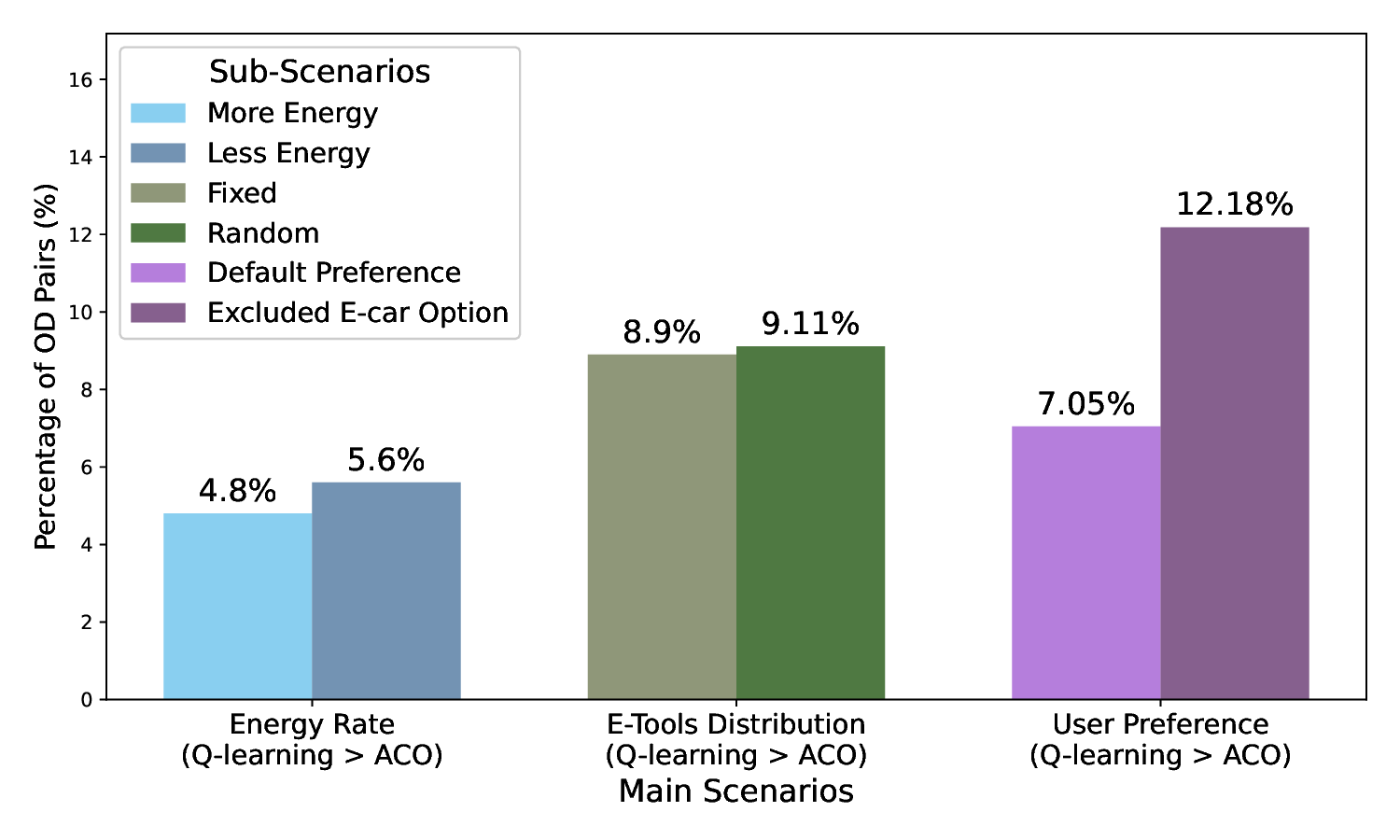}
        \caption{Bar Plot showing the Comparative Time Cost of Q-learning against MMEC-ACO corresponding to 3 different simulation aspects}
        \label{fig:multiple_scenarios}
\end{figure}

\textbf{Performance analysis for diverse simulation scenarios:} To emphasize the platform's ability to generate different scenarios in the context of shared E-mobility, we assess the performance of MMEC-ACO and Q-learning for different levels of initial SOC available to the E-mobility tools at different stations, user-preferences and distribution/placement of various E-mobility tools across the E-hubs/stations.
To illustrate the comparative results of the two algorithms and demonstrate the platform's ability to generate and manage different scenarios efficiently, we compiled and summarized the results for various scenarios. Due to space limitations, we do not present the raw results obtained for each algorithm. Instead, we present the overall statistics as a bar plot, as shown in Fig. \ref{fig:multiple_scenarios}. 

\textbf{Variations in SOC:} We assigned two different energy levels (50\% and 100\%) to the deployed tools and ran the algorithm for 500 (O, D) pairs for both MMEC-ACO and Q-learning (all other settings are fixed). The y-axis in Fig. \ref{fig:multiple_scenarios} shows the percentage of instances where Q-learning yields a larger optimal cost than MMEC-ACO. First two bars therefore indicate that Q-learning outperforms MMEC-ACO for 95.2\% of (O, D) pairs with low SOC levels and 94.4\% with high SOC levels, indicating its consistent superior performance.

\textbf{Variation in Distribution of E-mobility Tools:} The same trend is observed when changing the distribution pattern of E-mobility tools across E-hubs. In scenarios with fixed and random distributions of E-mobility tools, Q-learning outperforms MMEC-ACO for most origin-destination pairs. ``Fixed" distribution means each station consistently offers three E-mobility options, whereas ``random" distribution means these three options are allocated randomly but consistently across experiments. Q-learning performs better for 91.1\% of pairs in fixed setups and 90.89\% in random setups.

\textbf{User Preference:} This scenario addresses user preferences for certain E-mobility tools. The first bar represents equal preference for all tools, while the second bar pertains to scenarios where users do not prefer E-cars. Q-learning performs better for 92.95\% and 87.82\% of (O, D) pairs for equal preference and E-car excluded preference, respectively.

\section{Conclusion and Future Work} \label{conc}

In this paper, we presented an integrated open-source shared E-mobility platform to address the critical gap in E-mobility research. This platform offers a flexible and customizable architecture for various applications pertaining shared E-mobility. We showcased the platform's basic features and described its core modules. Additionally, we have demonstrated the platform's capabilities by employing the optimization module to provide optimal multi-modal route recommendations. This was achieved by utilizing the simulation module to generate background scenarios and relevant data for the optimization problem formulation. 
We used modified MMEC-ACO and Q-learning algorithms and provided a detailed comparison between the two for different scenarios. Findings indicate better performance of Q-learning as compared to ACO for  around 90\% of the tested (O, D) pairs amidst the variation in scenarios for energy availability, user-preference and distribution/placement of E-mobility tools. Direct comparison of runtime complexities for a fixed (O, D) pair shows Q-learning outperforming MMEC-ACO. For example, at around 2-second runtime, Q-learning achieves an optimal time cost of 3600 seconds, whereas MMEC-ACO achieves around 4500 seconds for a similar runtime.  Although we presented the overall architecture and a use-case demonstration, several aspects of the optimization process require further elaboration and will be addressed in a future extended version of this paper. Specifically, we will explore multi-user scenarios involving resource balancing and dynamic scheduling in greater detail.
Future work will also extend the discussion to include different optimization objectives, algorithm scalability concerning map size, the impact of the number of E-hubs on performance, the influence of vehicle numbers and traffic patterns, and a more detailed comparison with additional meta-heuristic and reinforcement-based optimization algorithms. 

\section*{ACKNOWLEDGMENT}

This research was supported by Science Foundation Ireland (Grant Numbers \textit{21/FFP-P/10266} and SFI/12/RC/2289P2) and the European Regional Development Fund, in collaboration with the SFI Insight Centre for Data Analytics at Dublin City University. Yue Ding was also supported by the SFI Centre for Research Training in Machine Learning (ML-Labs) at Dublin City University. We would like to extend our gratitude to Joe Naoum-Sawaya for his advice on problem formulation, and Conor McCarthy and Kevin O'Shea for their support with visualization and back-end implementation.

\bibliographystyle{splncs04}
\bibliography{references}

\begin{thebibliography}{10}
\providecommand{\url}[1]{\texttt{#1}}
\providecommand{\urlprefix}{URL }
\providecommand{\doi}[1]{https://doi.org/#1}

\bibitem{barcelo2005dynamic}
Barcel{\'o}, J., Casas, J.: Dynamic network simulation with aimsun. Simulation
  approaches in transportation analysis: Recent advances and challenges pp.
  57--98 (2005)

\bibitem{bird2023}
Bird: \url{https://www.bird.co/}, (accessed 23.04.2023)

\bibitem{buchmann2021stimulating}
Buchmann, T., Wolf, P., Fidaschek, S.: Stimulating e-mobility diffusion in
  germany (emosim): An agent-based simulation approach. Energies
  \textbf{14}(3), ~656 (2021)

\bibitem{Burani2022}
Burani, E., Cabri, G., Leoncini, M.: An algorithm to predict e-bike power
  consumption based on planned routes. Electronics  \textbf{11}(7), ~1105 (Mar
  2022). \doi{10.3390/electronics11071105},
  \url{http://dx.doi.org/10.3390/electronics11071105}

\bibitem{campisi2022promotion}
Campisi, T., Ali, N., Alemdar, K.D., Kaya, {\"O}., {\c{C}}odur, M.Y.,
  Tesoriere, G.: Promotion of e-mobility and its main share market: Some
  considerations about e-shared mobility. In: AIP Conference Proceedings.
  vol.~2611, p. 060002. AIP Publishing LLC (2022)

\bibitem{CSO2022Table}
{Central Statistics Office}: {Table SAP2022T11T1CTY - Census 2022 Preliminary
  Results}. Online (2022), \url{https://data.cso.ie/table/SAP2022T11T1CTY},
  accessed: 2023-06-27

\bibitem{chen2021review}
Chen, Y., Wu, G., Sun, R., Dubey, A., Laszka, A., Pugliese, P.: A review and
  outlook on energy consumption estimation models for electric vehicles. SAE
  International Journal of Sustainable Transportation, Energy, Environment, \&
  Policy  \textbf{2} (03 2021). \doi{10.4271/13-02-01-0005}

\bibitem{cho2022digital}
Cho, K.S., Park, S.W., Son, S.Y.: Digital twin-based simulation platform with
  integrated e-mobility and distribution system. In: CIRED Porto Workshop 2022:
  E-mobility and power distribution systems. vol.~2022, pp. 1158--1162. IET
  (2022)

\bibitem{DEANDRADE2018141}
{de Andrade}, R.C., Saraiva, R.D.: An integer linear programming model for the
  constrained shortest path tour problem. Electronic Notes in Discrete
  Mathematics  \textbf{69},  141--148 (2018).
  \doi{https://doi.org/10.1016/j.endm.2018.07.019},
  \url{https://www.sciencedirect.com/science/article/pii/S157106531830163X},
  joint EURO/ALIO International Conference 2018 on Applied Combinatorial
  Optimization (EURO/ALIO 2018)

\bibitem{dias2021therole}
Dias, G., Arsenio, E., Ribeiro, P.: The role of shared e-scooter systems in
  urban sustainability and resilience during the covid-19 mobility
  restrictions. Sustainability  \textbf{13}(13) (2021).
  \doi{10.3390/su13137084}, \url{https://www.mdpi.com/2071-1050/13/13/7084}

\bibitem{Dijkstra1959}
Dijkstra, E.W.: A note on two problems in connexion with graphs. Numerische
  Mathematik  \textbf{1}(1),  269–271 (Dec 1959). \doi{10.1007/bf01386390},
  \url{http://dx.doi.org/10.1007/BF01386390}

\bibitem{dorigo2006ant}
Dorigo, M., Birattari, M., Stutzle, T.: Ant colony optimization. IEEE
  computational intelligence magazine  \textbf{1}(4),  28--39 (2006)

\bibitem{duraismay2021adpaptive}
Duraisamy, T., Kaliyaperumal, D.: Adaptive passive balancing in battery
  management system for e‐mobility. International Journal of Energy Research
  \textbf{45} (02 2021). \doi{10.1002/er.6560}

\bibitem{echternacht2018simulating}
Echternacht, D., El~Haouati, I., Schermuly, R., Meyer, F.: Simulating the
  impact of e-mobility charging infrastructure on urban low-voltage networks.
  In: NEIS 2018; Conference on Sustainable Energy Supply and Energy Storage
  Systems. pp.~1--6. VDE (2018)

\bibitem{electricfeel2023}
ElectricFeel: \url{https://www.electricfeel.com/}, (accessed: 08.05.2023)

\bibitem{fellendorf2010microscopic}
Fellendorf, M., Vortisch, P.: Microscopic traffic flow simulator vissim.
  Fundamentals of traffic simulation pp. 63--93 (2010)

\bibitem{ferrara2019simulation}
Ferrara, M., Monechi, B., Valenti, G., Liberto, C., Nigro, M., Biazzo, I.: A
  simulation tool for energy management of e-mobility in urban areas. In: 2019
  6th International Conference on Models and Technologies for Intelligent
  Transportation Systems (MT-ITS). pp.~1--7. IEEE (2019)

\bibitem{DCC_map_ref}
Guériau, M., Dusparic, I.: Quantifying the impact of connected and autonomous
  vehicles on traffic efficiency and safety in mixed traffic. In: 2020 IEEE
  23rd International Conference on Intelligent Transportation Systems (ITSC).
  pp.~1--8 (2020). \doi{10.1109/ITSC45102.2020.9294174}

\bibitem{haklay2008openstreetmap}
Haklay, M., Weber, P.: Openstreetmap: User-generated street maps. IEEE
  Pervasive computing  \textbf{7}(4),  12--18 (2008)

\bibitem{krajzewicz2002sumo}
Krajzewicz, D., Hertkorn, G., R{\"o}ssel, C., Wagner, P.: Sumo (simulation of
  urban mobility)-an open-source traffic simulation. In: Proceedings of the 4th
  middle East Symposium on Simulation and Modelling (MESM20002). pp. 183--187
  (2002)

\bibitem{laurischkat2016business}
Laurischkat, K., Viertelhausen, A., Jandt, D.: Business models for electric
  mobility. Procedia Cirp  \textbf{47},  483--488 (2016)

\bibitem{liao2022electric}
Liao, F., Correia, G.: Electric carsharing and micromobility: A literature
  review on their usage pattern, demand, and potential impacts. International
  Journal of Sustainable Transportation  \textbf{16}(3),  269--286 (2022)

\bibitem{Liu_ACO}
Liu, M., Naoum-Sawaya, J., Gu, Y., Lecue, F., Shorten, R.: A distributed
  markovian parking assist system. IEEE Transactions on Intelligent
  Transportation Systems  \textbf{20}(6),  2230--2240 (2019).
  \doi{10.1109/TITS.2018.2865648}

\bibitem{drones6110365}
Liu, Y., Yan, S., Zhao, Y., Song, C., Li, F.: Improved dyna-q: A reinforcement
  learning method focused via heuristic graph for agv path planning in dynamic
  environments. Drones  \textbf{6}(11) (2022),
  \url{https://www.mdpi.com/2504-446X/6/11/365}

\bibitem{luo2021deployment}
Luo, M., Du, B., Klemmer, K., Zhu, H., Wen, H.: Deployment optimization for
  shared e-mobility systems with multi-agent deep neural search. IEEE
  Transactions on Intelligent Transportation Systems  \textbf{23}(3),
  2549--2560 (2021)

\bibitem{luo2023fleet}
Luo, M., Du, B., Zhang, W., Song, T., Li, K., Zhu, H., Birkin, M., Wen, H.:
  Fleet rebalancing for expanding shared e-mobility systems: A multi-agent deep
  reinforcement learning approach. IEEE Transactions on Intelligent
  Transportation Systems  (2023)

\bibitem{wondermobility2023}
Mobility, W.: \url{https://www.wundermobility.com/}, (accessed: 08.05.2023)

\bibitem{moovit2023}
Moovit: \url{https://moovitapp.com/sf_bay_area_ca-22/poi/en-gb}, (accessed:
  08.05.2023)

\bibitem{Musa2023}
Musa, A.A., Malami, S.I., Alanazi, F., Ounaies, W., Alshammari, M., Haruna,
  S.I.: Sustainable traffic management for smart cities using
  internet-of-things-oriented intelligent transportation systems (its):
  Challenges and recommendations. Sustainability  \textbf{15}(13), ~9859 (Jun
  2023). \doi{10.3390/su15139859}, \url{http://dx.doi.org/10.3390/su15139859}

\bibitem{nikitas2021cycling}
Nikitas, A., Tsigdinos, S., Karolemeas, C., Kourmpa, E., Bakogiannis, E.:
  Cycling in the era of covid-19: Lessons learnt and best practice policy
  recommendations for a more bike-centric future. Sustainability  \textbf{13},
  ~4620 (04 2021). \doi{10.3390/su13094620}

\bibitem{Watkins1992}
Watkins, C.J.C.H., Dayan, P.: Q-learning. Machine Learning  \textbf{8}(3–4),
  279–292 (May 1992). \doi{10.1007/bf00992698},
  \url{http://dx.doi.org/10.1007/BF00992698}

\bibitem{wu2017flow}
Wu, C., Kreidieh, A., Parvate, K., Vinitsky, E., Bayen, A.M.: Flow:
  Architecture and benchmarking for reinforcement learning in traffic control.
  arXiv preprint arXiv:1710.05465  \textbf{10} (2017)

\bibitem{Sen_Fed}
Yan, S., Fang, H., Li, J., Ward, T., O’Connor, N., Liu, M.: Privacy-aware
  energy consumption modeling of connected battery electric vehicles using
  federated learning. IEEE Transactions on Transportation Electrification
  (2023). \doi{10.1109/TTE.2023.3343106}

\bibitem{energy_review}
Yan, S., Shah, M.H., Li, J., O’Connor, N., Liu, M.: A review on ai algorithms
  for energy management in e-mobility services. In: 2023 7th CAA International
  Conference on Vehicular Control and Intelligence (CVCI). pp.~1--8 (2023).
  \doi{10.1109/CVCI59596.2023.10397371}

\bibitem{ding2024datadriven}
Yue, D., Sen, Y., Shah, M.H., Hongyuan, F., Ji, L., Liu, M.: Data-driven energy
  consumption modelling for electric micromobility using an open dataset
  (2024), accepted by the 2024 IEEE Transportation Electrification Conference
  and Expo

\end{thebibliography}

\end{document}